\pdfoutput=1

\documentclass[11pt]{article}

\usepackage[final]{acl}

\usepackage{times}
\usepackage{latexsym}

\usepackage[T1]{fontenc}

\usepackage[utf8]{inputenc}

\usepackage{microtype}

\usepackage{inconsolata}

\usepackage{graphicx}

\newcommand{\mname}{{MiLoRA}\xspace}
\newcommand{\mnamec}{MiLoRA}

\usepackage{amsmath}
\usepackage{amsfonts}
\usepackage{algorithm}
\usepackage{algorithmic}
\usepackage[algo2e]{algorithm2e}
\usepackage{graphicx}
\usepackage{multirow}
\usepackage{subfigure}
\usepackage{subcaption}
\usepackage{caption}
\usepackage{wrapfig}
\usepackage{booktabs}

\usepackage{enumitem}
\setlist[itemize]{align=parleft,left=0pt..1em}

\usepackage{amsmath,amsfonts,bm}

\def\eqref#1{equation~\ref{#1}}

\def\1{\bm{1}}

\def\mA{{\bm{A}}}
\def\mB{{\bm{B}}}

\def\mU{{\bm{U}}}
\def\mV{{\bm{V}}}
\def\mW{{\bm{W}}}

\def\mSigma{{\bm{\Sigma}}}

\DeclareMathAlphabet{\mathsfit}{\encodingdefault}{\sfdefault}{m}{sl}
\SetMathAlphabet{\mathsfit}{bold}{\encodingdefault}{\sfdefault}{bx}{n}

\title{MiLoRA: Harnessing Minor Singular Components for \\ Parameter-Efficient LLM Finetuning}

\author{
Hanqing Wang$^{1}$, \
Yixia Li$^{2}$, \
Shuo Wang$^{3}$, \
Guanhua Chen$^{2}$\thanks{Corresponding Authors.}\;, \
Yun Chen$^{1,4}$\footnotemark[1] \\
$^1$Shanghai University of Finance and Economics  \\
$^2$Southern University of Science and Technology,\; $^3$Tsinghua University \\ 
$^4$MoE Key Laboratory of Interdisciplinary Research of Computation and Economics, \\
Shanghai University of Finance and Economics \\
}

\begin{document}
\maketitle
\begin{abstract}
Efficient finetuning of large language models (LLMs) aims to adapt the LLMs with reduced computational and memory costs. Previous LoRA-based approaches initialize the low-rank matrices with Gaussian distribution and zero values while keeping the original weight matrices frozen. However, the trainable model parameters optimized in an unguided subspace might interfere with the well-learned subspace of the pretrained weight matrices.
In this paper, we propose MiLoRA, a simple yet effective LLM finetuning approach that only updates the minor singular components of the weight matrix while keeping the principal singular components frozen. It is observed that the minor matrix corresponds to the noisy or long-tail information, while the principal matrix contains important knowledge. The MiLoRA initializes the low-rank matrices within a subspace that is orthogonal to the principal matrix, thus the pretrained knowledge is expected to be well preserved. During finetuning, MiLoRA makes the most use of the less-optimized subspace for learning the labeled dataset. Extensive experiments on commonsense reasoning, math reasoning, instruction following and visual instruction following benchmarks present the superior performance of our method.
\end{abstract}

\section{Introduction}

Large language models \citep[LLMs]{brown2020gpt3,ouyang2022instructgpt,touvron2023llama2,jiang2023mistral} have demonstrated superior performance on various tasks \citep{zheng2023judging}, such as math reasoning \citep{wang2024mathshepherd} and question answering \citep{ivison2023camels}. These models are pretrained with the next token prediction task \citep{brown2020gpt3} on large web-scale data, then finetuned with instruction data as well as human preference data \citep{ouyang2022instructgpt,yu2023metamath,cui2023ultrafeedback} for different downstream tasks.
Fully finetuning is commonly employed to unlock the complete potential of LLMs, however, optimizing all model parameters necessitates substantial and restrictive computational resources \citep{touvron2023llama2,jiang2023mistral}, which hampers the utilization of LLMs across diverse scenarios.

Parameter-efficient finetuning \citep[PEFT]{hu2021lora,hu2023llmadapters,liu2022ptuning,zhang2024llamaadapter,liu2024dora} aims at reducing the computational and GPU memory cost for finetuning of pretrained models. Low-rank adaptation \citep[LoRA]{hu2021lora} is one of the most widely used PEFT methods for LLM finetuning. It assumes the change of linear model weights to be low-rank \citep{li2018measuring,aghajanyan-etal-2021-intrinsic}. For each selected weight matrix, it only updates two low-rank matrices while keeping the pretrained weight frozen. During inference, the low-rank matrices are merged into the pretrained linear weights, thus no additional computational or memory cost is introduced.
Recently, researchers have explored different LoRA-based variants for efficient LLM finetuning \citep{zhang2023adalora,pan2024lisa,kopiczko2024vera,liu2024dora,meng2024pissa}. However, most existing LoRA-based methods randomly initialize the low-rank matrices and optimize the trainable parameters in an unguided subspace. We suspect this strategy may override important pretrained features, thus degrading the performance of low-rank adaptation methods \citep{dou2024loramoe}.

In this paper, we propose \textbf{Mi}nor singular component based \textbf{Lo}w \textbf{R}ank \textbf{A}daptation (\mname) for efficient LLM finetuning. \mname has a similar framework as LoRA but employs a different initialization schedule. Specifically, a weight matrix $\mW$ is decomposed with the singular value decomposition (SVD) algorithm. Based on the magnitude of the singular values, we divide $\mW$ into two components: the principal matrix $\mW_p$ corresponding to large singular values and the minor matrix $\mW_m$ corresponding to small singular values.
We argue that the principal matrix captures the essence of the pretrained knowledge, whereas the minor matrix is suboptimal with noisy or long-tail information.
It is supported by previous works \citep{Hajimolahoseini2021CompressingPL,sharma2024laser,NEURIPS2024_8555cf30} that principal low-rank approximation can achieve comparable or even better performance to full parameters.

Motivated by these observations, we freeze the principal matrix $\mW_p$ and adapt the minor singular components during finetuning. The low-rank matrices $\mA$ and $\mB$ in LoRA framework are initialized with the minor matrix $\mW_m$.
Since the trainable low-rank matrices are initialized in a subspace orthogonal to the principal matrix, \mname is expected to effectively learn from finetuning tasks while better preserving and utilizing the pretrained knowledge.
To maintain the capability of the pretrained model at the start of finetuning, vanilla LoRA explicitly initializes $\mB$ with zeros. In contrast, \mname naturally satisfies this requirement, as the pretrained weight matrix $\mW$ equals the frozen principal part $\mW_p$ plus the low-rank part $\mW_m=\mB \mA$.
We conduct extensive experiments on commonsense reasoning, math reasoning, instruction following and visual instruction following benchmarks. The experimental results show that \mname consistently outperforms LoRA without sacrificing training or inference efficiency, such
as commonsense reasoning (+1.6/+1.1 on LLaMA2-7B/LLaMA3-8B), math reasoning (+2.0 on LLaMA2-7B), instruction following (+2.9 on LLaMA2-7B) and visual instruction following (+1.4 on LLaVA1.5-7B).\footnote{Our code and model are publicly available at \url{https://github.com/sufenlp/MiLoRA}.}

\section{Preliminaries}
\paragraph{Singular Value Decomposition}
Given a matrix $\mW \in \mathbb{R}^{m \times n}$, its singular value decomposition is denoted as $\mW=\mU \bm{\Sigma} \mV^{\top}$, where $\mU=\left[u_1, u_2, \cdots, u_m\right] \in \mathbb{R}^{m \times m}$, $\mV=\left[v_1, v_2, \cdots, v_n\right] \in \mathbb{R}^{n \times n}$. The columns of $\mU$ are the left singular vectors, and the columns of $\mV$ are the right singular vectors. The $\bm{\Sigma} \in \mathbb{R}^{m \times n}$ is a diagonal matrix containing the singular values of $\mW$ in descending order.
Without loss of generality, we suppose $m \leq n$ to simplify the notation. The SVD of $\mW$ can be reformulated as
\begin{align}
\mW = \mU \bm{\Sigma} \mV^{\top} = \sum_{i=1}^{m} \sigma_i u_i v_i^{\top},  \label{eq:svd-decomp}
\end{align}
where $u_i$ and $v_i$ are the $i^{\mathrm th}$ column of $\mU$ and $\mV$, respectively.

\paragraph{Low-Rank Adaptation}
The low-rank adaptation method \citep[LoRA]{hu2021lora} assumes the updates of linear weight $\mW \in \mathbb{R}^{m \times n}$ to be low-rank, thus models the changes with two trainable low-rank matrices $\mA \in \mathbb{R}^{r\times n}$ and $\mB \in \mathbb{R}^{m\times r}$. The weight matrix can be decomposed as
\begin{align}
    \mW = \mW^{(0)} + \Delta \mW = \mW^{(0)} + \frac{\alpha}{r} \mB \mA, \label{eq:lora}
\end{align}
where $\mW^{(0)}$ and $\Delta \mW$ refer to the pretrained weight and weight change, respectively. The $\alpha$ and $r$ are hyperparameters of scaling-factor and LoRA rank ($r \ll \min(m,n)$).
During finetuning, the pretrained matrix $\mW^{(0)}$ is kept frozen. It significantly diminishes the number of trainable parameters as both $\mA$ and $\mB$ matrices are low-rank.
The $\mB$ matrix is initialized with zero while $\mA$ matrix adopts a random Gaussian distribution with zero mean value.
This initialization strategy ensures the $\Delta \mW =0$ at the beginning of training. The LoRA method only modifies the linear matrices in the Transformer model. The low-rank matrices can be easily merged into the pretrained linear matrix to get updated for inference, which does not require additional computing and GPU memory compared with full finetuning.
However, the vanilla LoRA method fails to select the optimal subspace for updating the model parameters, as the low-rank matrices $\mA$ and $\mB$ are randomly initialized. This might potentially detract from the pretrained knowledge encoded in the pretrained weight matrix.

\begin{figure*}[t]
    \centering
    \includegraphics[width=1.5\columnwidth]{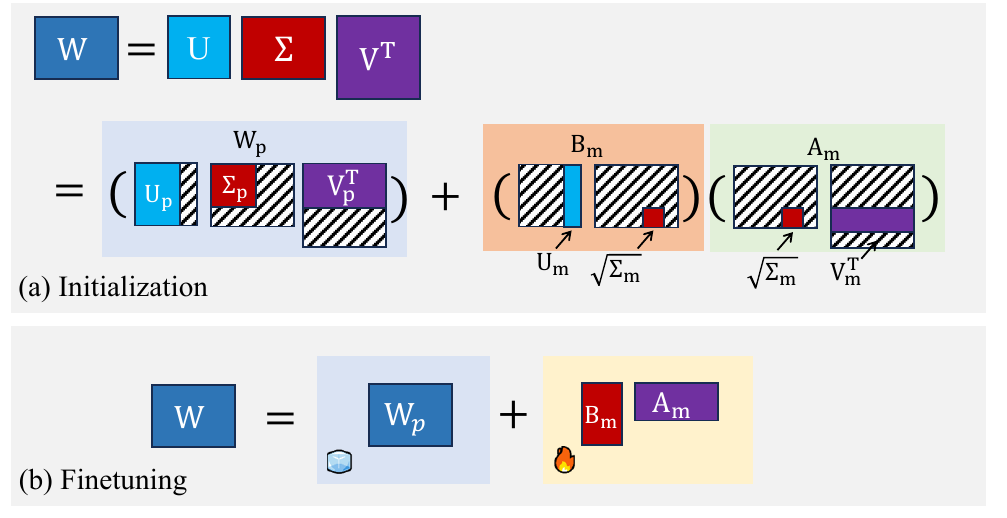}
    \caption{(a) \mname method splits the pretrained linear weight matrix into two parts, the minor singular component ($\mU_m, \mV_m, {\bm \Sigma}_m$) is used to initialize the low-rank matrices $\mA_m$ and $\mB_m$ in the LoRA framework. (b) During finetuning, only the low-rank matrices $\mA_m$ and $\mB_m$ are updated while the principal matrix $\mW_p$ is frozen, which is similar to the practice of LoRA method. }
    \label{fig:milora-overview}
    \vspace{10pt}
\end{figure*}

\section{Methodology}
The proposed \mname is a simple yet effective PEFT approach that selects the minor singular components of the weight matrices for optimization. As observed in LASER \citep{sharma2024laser}, the minor singular components of weight matrices contain noisy or long-tail information, while the principal singular components are responsible for important features across tasks. Therefore, \mname is designed to effectively learn from the finetuning dataset by adapting minor singular components, while maintaining the knowledge encoded in the pretrained model. On the contrary, vanilla LoRA fails to constrain the optimization subspace as its low-rank matrices are randomly initialized.
Specifically, suppose $m \leq n$, \mname splits each linear weight matrix into two matrices according to the corresponding singular values: principal matrix $\mW_{p}$ and minor matrix $\mW_{m}$ (see Figure~\ref{fig:milora-overview}):
\begin{equation}
\begin{aligned}
\mW &=\mW_{p} + \mW_{m} \\
&= \sum_{i=1}^{m-r} \sigma_i u_i v_i^{\top} + \sum_{i=m-r+1}^{m} \sigma_i u_i v_i^{\top} ,  \label{eq:milora}
\end{aligned}
\end{equation}
where the singular values are in descending order ($\sigma_1 \geq \sigma_2 \geq \cdots \geq \sigma_m \geq 0$), and the hyperparameter $r$ is the number of minor singular values in the $\mW_m$ matrix.

The decomposition can also be reformulated in a matrix form. The $\mU$ matrix in the SVD of $\mW$ can be reformulated as $\mU=\left[\mU_p, \mU_m\right]$, where $\mU_p = \left[u_1, u_2, \cdots, u_{m-r} \right]$ and $\mU_m=\left[u_{m-r+1}, u_{m-r+2}, \cdots, u_{m} \right]$ are left singular vectors corresponding to principal and minor singular values, respectively. The $\mV$ and $\mSigma$ matrices can be reformulated similarly. The SVD of $\mW$ can be expressed as:
\begin{equation}
\begin{aligned}
\mW &= \mU \mSigma \mV^{\top}
= \mU_p \mSigma_p \mV_p^{\top} + \mU_m \mSigma_m \mV_m^{\top} \\
&= \mW_p + \mW_m
\end{aligned}
\end{equation}

During the finetuning process, instead of freezing the entire pretrained weight matrix, we just keep the principal singular components $\mW_{p}$ fixed to preserve the pretrained knowledge. The minor matrix $\mW_m$ is used to initialize low-rank matrices $\mA_m$ and $\mB_m$ (see Figure~\ref{fig:milora-overview}):
\begin{equation}
\begin{aligned}
\mW_m &=\mU_m \mSigma_m \mV_m^{\top}\\
&=(\mU_m \sqrt{\mSigma_m}) (\sqrt{\mSigma_m} \mV_m^{\top})=\mB_m \mA_m,
\end{aligned}
\end{equation}

This strategy has two benefits: 1) It encourages the model to learn in the less-optimized subspace spanned by the minor singular vectors, thus reducing the interference with the well-learned pretrained knowledge encoded in the principal singular components. 2) Unlike vanilla LoRA which requires tuning the scaling factor $\alpha$ and the initialization hyperparameters, ours does not require any hyperparameter except the rank $r$.

One potential variant of \mname is to impose strict constraints on the orthogonality between $\mW_m$ and $\mW_p$ during fine-tuning. However, we adopt a softer approach by guiding the optimization direction solely through initialization, resulting in a method that is simpler, more training-efficient, and more flexible. Our experiment results show that our scheme works quite well in balancing learning from the finetuning dataset and preserving pretrained knowledge. 

\section{Experiments}\label{sec:expriments}

\begin{table*}[t]
\centering
\resizebox{1.0\textwidth}{!}
{
\begin{tabular}{lcccccccccc}
\toprule
Model& PEFT & BoolQ & PIQA & SIQA & HellaSwag & WinoGrande & ARC-e & ARC-c & OBQA & Avg. \\
\midrule
ChatGPT & $-$ & 73.1 & 85.4 & 68.5 & 78.5 & 66.1 & 89.8 & 79.9 & 74.8 & 77.0 \\
\midrule
\multirow{3}{*}{LLaMA2-7B} & LoRA$^\dagger$ & \textbf{69.8} & 79.9 & 79.5 & 83.6 & \textbf{82.6} & 79.8 & 64.7 & \textbf{81.0} & 77.6 \\
& PiSSA & 67.6 & 78.1 & 78.4 & 76.6 & 78.0 & 75.8 & 60.2 & 75.6 & 73.8 \\
& \mname & 67.6 & \textbf{83.8} & \textbf{80.1} & \textbf{88.2} & 82.0 & \textbf{82.8} & \textbf{68.8} & 80.6 & \textbf{79.2} \\
\midrule
\multirow{3}{*}{LLaMA3-8B} & LoRA$^\dagger$ & \textbf{70.8} & 85.2 & \textbf{79.9} & 91.7 & 84.3 & 84.2 & 71.2 & 79.0 & 80.8 \\
& PiSSA & 67.1 & 81.1 & 77.2 & 83.6 & 78.9 & 77.7 & 63.2 & 74.6 & 75.4 \\
& \mname & 68.8 & \textbf{86.7 }& 77.2 & \textbf{92.9} & \textbf{85.6} & \textbf{86.8} & \textbf{75.5} & \textbf{81.8} & \textbf{81.9} \\
\bottomrule
\end{tabular}
}
\caption{Commonsense reasoning evaluation results for LLaMA2-7B and LLaMA3-8B on eight tasks. The reported metric in this table is accuracy. $^\dagger$Results are cited from \citet{liu2024dora}. Bold numbers indicate the highest performance scores for each dataset across the different PEFT methods for the corresponding model.}
\label{tab:commonsense}
\end{table*}

\begin{table}[t]
\centering
\begin{tabular}{cccc}
\toprule
Method & GSM8K & MATH &Avg. \\
\midrule
Full FT$^\dagger$ & 66.5 & 19.8 & 43.2 \\ \midrule
LoRA & 60.6 & 16.9 & 38.7 \\
PiSSA & 58.2 & 15.8 & 37.0 \\
\mname & \textbf{63.5} & \textbf{17.8} & \textbf{40.7} \\
\bottomrule
\end{tabular}
\caption{Math reasoning evaluation results for LLaMA2-7B on GSM8K and MATH. The evaluation metric presented in this table is the Exact Match (EM) ratio. $^\dagger$Results are cited from \citet{yu2023metamath}. For each dataset, the highest performance scores among the different PEFT methods are shown in bold.}
\label{tab:math}
\end{table}

To evaluate \mname against other PEFTs, we conduct experiments on both NLP and multimodal tasks covering 18 datasets. We benchmark against LoRA \citep{hu2021lora} and PiSSA \citep{meng2024pissa}. All of our experiments are conducted on 8 NVIDIA L40 GPUs unless otherwise specified.
\begin{itemize}
    \item LoRA \citep{hu2021lora} reparameterizes the weight update $\Delta \mW$ with two trainable low-rank matrices $\mA$ and $\mB$, while freezing the pretrained weight $\mW$. They use a random Gaussian initialization for A and zero for B, so $\Delta \mW = \mB \mA$ is zero at the beginning of training.
    \item PiSSA \citep{meng2024pissa} shares the same framework as LoRA, but employs a more sophisticated initialization approach. They initialize $\mA$ and $\mB$ with principal singular values and singular vectors of the pre-trained weight $\mW$.
    Given that the principal components capture the essence of a matrix, PiSSA is expected to better approximate full finetuning by changing the essential parts while freezing the ``noisy'' parts.
\end{itemize}

\subsection{Experiments on Large Language Model}\label{exp:llm}
\paragraph{Models and Datasets} We compare \mname with baselines on three different types of LLM downstream tasks. 

\noindent$\bullet\,$\textbf{Commonsense reasoning:} We finetune LLaMA2-7B \citep{touvron2023llama2} and LLaMA3-7B \citep{llama3modelcard} on Commonsense170K \citep{hu2023llmadapters}.
Eight commonsense reasoning datasets are used for evaluation, including BoolQ~\citep{clark-etal-2019-boolq}, PIQA~\citep{bisk2020piqa}, SIQA~\citep{sap-etal-2019-social}, HellaSwag~\citep{zellers2019hellaswag}, WinoGrande~\citep{sakaguchi2021winogrande}, ARC-e, ARC-c~\citep{clark2018think}, and OBQA~\citep{mihaylov2018can}. The task is formulated as a multiple-choice problem. We report accuracy (\%) for all datasets on the best checkpoint chosen by the validation set loss.

\noindent $\bullet\,$\textbf{Math reasoning:} We finetune LLaMA2-7B~\citep{touvron2023llama2}  on the MetaMathQA dataset \citep{yu2023metamath}, which contains 395K samples augmented from the training set of GSM8K~\citep{cobbe2021training} and MATH~\citep{hendrycksmath2021}.
We use test sets of GSM8K and MATH for evaluation and report results on the last checkpoint.
We report the Exact Match (EM) ratio against the ground truth for each test set.

\noindent $\bullet\,$\textbf{Instruction following:} We finetune LLaMA2-7B with Ultrafeedback~ \citep{cui2023ultrafeedback} following previous works \cite{wu2024advancing,wu2024reft}. We evaluate our models using AlpacaEval 2.0~\citep{dubois2024length,alpaca_eval}, FollowBench~\citep{jiang-etal-2024-followbench}, and IFEval~\citep{zhou2023instruction}. For AlpacaEval 2.0, we use Length-controlled Win Rate (LC WR) and Win Rate (WR) against GPT-4 Turbo~\citep{OpenAI2023} as evaluation metrics. In FollowBench, we adopt Hard Satisfaction Rate (HSR), while in IFEval, we measure accuracy.

\begin{table*}[t]
\centering
\resizebox{1.0\textwidth}{!}
{
\begin{tabular}{l|cc|ccccc|cccc|c}
\toprule
\multirow{2}{*}{Method} & \multicolumn{2}{c|}{AlpacaEval 2.0} & \multicolumn{5}{c|}{FollowBench}  &  \multicolumn{4}{c|}{IFEval} & \multirow{2}{*}{Avg.} \\
\cmidrule(lr){2-3} \cmidrule(lr){4-8} \cmidrule(lr){9-12} 
& LC WR & WR & Level1 & Level2 & Level3 & Level4 &  Level5 & Prompt-strict & Ins-strict & Prompt-loose & Ins-loose \\
\midrule
LoRA & 5.6          & 3.9 & \textbf{46.5} & \textbf{39.2} & 35.5          & 22.2          & 14.3             & 28.8          & 39.9          & 31.1          & 42.3   & 28.1       \\
PiSSA & 5.6          & 3.9 & 44.2          & 37.1          & 30.6          & 20.9          & 14.3            & 31.6          & 42.7          & 34.6          & 45.7  & 28.3        \\
\mname & \textbf{7.1} & \textbf{4.4} & 45.1          & 37.6          & \textbf{38.2} & \textbf{28.5} & \textbf{20.8} & \textbf{32.4} & \textbf{44.8} & \textbf{34.8} & \textbf{47.7} & \textbf{31.0} \\
\bottomrule
\end{tabular}
}
\caption{Instruction following evaluation results for LLaMA2-7B on AlpacaEval 2.0, FollowBench, and IFEval. AlpacaEval 2.0 is evaluated using Length-controlled Win Rate (LC WR) and Win Rate (WR) against GPT-4 Turbo~\citep{OpenAI2023}, FollowBench adopts Hard Satisfaction Rate (HSR), and IFEval is measured by accuracy. Bold numbers indicate the highest performance scores for each dataset across the different PEFT methods for the corresponding model.}
\label{tab:instruction_following}
\vspace{10pt}
\end{table*}

\paragraph{Implementation Details}
We use the same hyperparameter configurations as~\citet{hu2023llmadapters} without tuning for all methods. Details can be found in Appendix~\ref{sec:app_hy}. We denote this hyperparameter setup as LLM-Adapters. By default, we use a rank of 32. As math reasoning has a large training dataset, we set the rank to 64. We use the implementation of LLM-Adapters \citep{hu2023llmadapters}\footnote{\url{https://github.com/AGI-Edgerunners/LLM-Adapters}} for commonsense reasoning, the implementation of PiSSA \citep{meng2024pissa}\footnote{\url{https://github.com/GraphPKU/PiSSA}} for math reasoning and the implementation of open-instruct\footnote{\url{https://github.com/allenai/open-instruct}} for instruction tuning.

\paragraph{Results}
\begin{table*}[t]
\centering
\vspace{-15pt}
{
\begin{tabular}{lcccccccc}
\toprule
Method  & VQAv2         & GQA           & VisWiz        & SQA           & VQAT          & POPE          & MMBench       & Avg.          \\
\midrule
Full FT$^\dagger$           & 78.5          & 61.9          & 50.0          & 66.8          & 58.2          & 85.9          & 64.3          & 66.5          \\
\midrule
LoRA$^\dagger$           & 79.1          & \textbf{62.9} & 47.8          & 68.4          & 58.2          & 86.4          & 66.1          & 66.9          \\
PiSSA              & 77.5              & 60.6              & 39.2              & 67.3              & 54.3              &  87.0             & 61.4              & 63.9              \\
\mname        & \textbf{79.2} & 62.1          & \textbf{53.3} & \textbf{70.6} & \textbf{58.7} & \textbf{87.9} & \textbf{66.1} & \textbf{68.3} \\
\bottomrule
\end{tabular}
}
\caption{Visual instruction tuning evaluation results for LLaVA1.5-7B on seven vision-language tasks. Accuracy serves as the metric reported in this table. $^\dagger$We directly use checkpoints from \citet{liu2024visual} to reproduce the results of baseline methods.}
\label{tab:visual_instruction_tuning}
\vspace{-5pt}
\end{table*}

We report results in Table~\ref{tab:commonsense} for commonsense reasoning, Table~\ref{tab:math} for math reasoning, and Tables~\ref{tab:instruction_following} for instruction following, respectively. For commonsense reasoning, we also include the ChatGPT baseline reported in \citet{liu2024dora} as a reference, which is obtained with GPT-3.5 Turbo~\citep{OpenAIGPT35Turbo2023} API using a zero-shot Chain of Thought \citep{wei2022chain}. For math reasoning, we add the full finetuning results from \citet{yu2023metamath} as a reference.

As can be seen, \mname consistently surpasses all baseline methods across different datasets and LLMs, indicating that \mname is a highly effective finetuning method. Specifically, for commonsense reasoning, \mname outperforms LoRA and PiSSA by an average of 1.6 and 5.4 accuracy points on LLaMA2-7B and 1.1 and 6.5 points on LLaMA3-8B. For math reasoning, \mname 
improves over LoRA and PiSSA by 2.0 and 3.7 averaged Exact Match (EM) points on LLaMA2-7B, respectively. However, it still underperforms full finetuning by an average of 2.5 EM score. This suggests that the PEFT methods still have room for improvement to fully match the performance of full fine-tuning. For instruction following, the results on AlpacaEval 2.0 indicate that PiSSA performs comparably to LoRA, while MiLoRA outperforms both by 1.5 in LC WR and 0.5 in WR, demonstrating its superior effectiveness. Furthermore, experiments on FollowBench and IFEval show that MiLoRA significantly surpasses both LoRA and PiSSA across these benchmarks. Notably, MiLoRA excels on more challenging problems (levels 4 and 5) in FollowBench, likely due to its ability to better retain pre-trained knowledge, which is crucial for handling complex instructions. Overall, MiLoRA surpasses LoRA and PiSSA in instruction following, achieving an average improvement of 2.9 and 2.7 points, respectively, further underscoring its advantages.

\subsection{Experiments on Vision-Language Model} \label{sec:vl_exp}

\paragraph{Models and Datasets}
To further examine if \mname can remain competitive on multimodal finetuning tasks, we further conduct visual instruction tuning tasks on LLaVA1.5-7B \cite{liu2024visual}, which consists of an LLM, Vicuna-1.5-7B \cite{peng2023instruction} and a vision encoder, CLIP ViT-L/336px \cite{Radford2021LearningTV}. The visual instruction tuning is performed in a multitask setting, including VQA \cite{hudson2019gqa,marino19okvqa,schwenk22aokvqa}, OCR \cite{ocrvqa,textcaps}, region-level VQA \cite{krishna17vg,mao16generation}, visual conversation \cite{liu2024visual}, and language conversation data. The finetuned models are evaluated on seven different multimodal benchmarks: VQA-v2 \cite{goyal2017making}, GQA \cite{hudson2019gqa}, VizWiz \cite{gurari2018vizwiz}, SQA \cite{lu22sqa}, VQAT \cite{singh19vqat}, POPE \cite{li2023pope}, and MMBench \cite{liu2024mmbench}.

\paragraph{Implementation Details}
We follow the settings of \citet{liu2024visual} to construct the training data and prompt template. For the hyperparameter configuration, we follow the LoRA configuration provided by \citet{liu2024visual} without tuning for all PEFT methods for a fair comparison. Details can be found in Appendix~\ref{sec:more_imple_details}.

\paragraph{Results}
Table \ref{tab:visual_instruction_tuning} shows the experiment results. As can be seen, \mname achieves the best results, outperforming LoRA and PiSSA by 1.4 and 4.4 average accuracy points, respectively. We also note that in this setup, fully fine-tuning is less effective than LoRA. As a result, methods like PiSSA, which aim to closely approximate fully fine-tuning, may lose their advantage. In contrast, \mname improves performance by retaining more of the pre-trained knowledge, thereby circumventing this issue.

\section{Understanding \mnamec}
In this section, we conduct experiments to further understand \mnamec. By default, we use the finetuned LLaMA2-7B model on math reasoning from our main experiments for analyses.

\subsection{Comparison with More PEFTs} \label{sec:more_results}

\begin{table}[t]
\centering
\resizebox{1.0\columnwidth}{!}{
\begin{tabular}{lccc}
\toprule
Method  & GSM8K          & Human-eval &  Training Cost\\
\midrule
rsLoRA  & 45.6          & 16.0   & $\times$ 1.00 \\
LoRA+   & 52.1          & 18.2   & $\times$ 1.00 \\
DoRA    & 53.1          & 19.8   & $\times$ 1.58 \\
AdaLoRA & 50.7          & 17.8   & $\times$ 1.07  \\
LoRA-GA & 53.6          & 19.8   & $\times$ 1.00 \\
\midrule
\mname                 & \textbf{54.7} &  \textbf{24.4}   & $\times$ 1.00      \\
\bottomrule
\end{tabular}
}
\caption{Comparison with more baselines on math and code tasks. Baseline GSM8K and Human-eval results are cited from \citet{wang2024loraga}. The evaluation metric for GSM8K is Exact Match (EM), while for Human-eval, it is Pass@1.}
\label{tab:more_comp_pefts}
\vspace{-5pt}
\end{table}

To further demonstrate the effectiveness of \mname, we compare \mname with additional popular baselines, such as rsLoRA \cite{kalajdzievski2023rslora}, LoRA+ \cite{hayou2024lora}, DoRA \cite{liu2024dora}, AdaLoRA \cite{zhang2023adalora}, LoRA-GA \cite{wang2024loraga}. We follow the math and code experiment settings in \citet{wang2024loraga}. Specifically, we finetune LLaMA2-7B with \mname on MetaMathQA 100K and Code-Feedback 100K for one epoch using rank $8$ and compare \mname with baseline results reported in \citet{wang2024loraga}. Table~\ref{tab:more_comp_pefts} shows the results. We report the training cost of all methods relative to LoRA, which requires 3.58 hours of training on the math task using a single NVIDIA-L40 GPU. As can be seen, \mname consistently outperforms all baselines, improving the best PEFT baseline LoRA-GA by 1.1 and 4.6 points on GSM8K and Human-eval, respectively. The training efficiency of \mname is also very high. The time for SVD decomposition in \mname is less than six minutes and negligible when compared with the total training time.
Therefore, the training time required of \mname is consistent with that of LoRA, while DoRA and AdaLoRA require additional training time overhead.

\subsection{Is Minor Singular Component Adaptation Important?}\label{sec:comp}

\begin{table}[t]
\centering
{
\begin{tabular}{l|ccc}
\toprule
Method & Principal & Random & Minor \\
\midrule
GSM8K & 60.7 & 63.2 & \textbf{64.0} \\
MATH & 14.6 & 15.5 & \textbf{16.1} \\
\bottomrule
\end{tabular}
}
\caption{Performance of \mname when initializing with principal, random sampled, and minor singular components.}
\label{tab:mag}
\vspace{-10pt}
\end{table}

To investigate the influence of singular components of varying magnitudes on finetuning performance, we conduct experiments on math reasoning tasks using the LLaMA2-7B model. Specifically, we finetune LLaMA2-7B on the MetaMathQA 395K dataset for 1 epoch with a maximum sequence length of 512 to save computation. We initialize the low-rank matrices $\mA$ and $\mB$ with principal, minor, and randomly sampled singular components, and report the evaluation results in Table \ref{tab:mag}. As can be seen, initializing with the minor singular components achieves the best performance across both datasets.
This underscores the importance of adapting the minor singular components during the finetuning process.

\begin{table*}[t]
\centering
\resizebox{0.65\textwidth}{!}{
\begin{tabular}{r|cc|ccc}
\toprule
$\left\|\mW\right\|_F=97.36$ & $\mW$ &  Random & LoRA & PiSSA & \mname \\
\midrule
$\left\|\mU^{\top} \mW \mV \right\|_F=$ & 43.64 & 1.50 & 1.82 & 17.57 & 1.86  \\
\midrule
$\left\|\mU^{\top} \Delta \mW \mV \right\|_F=$ & - & - & 68.18 & 55.79 & 44.95  \\
\midrule
Ampliﬁcation factor & - & - & 37.46 & 3.18 & 24.17 \\
\bottomrule
\end{tabular}}
\caption{The Frobenius norm of $\mU^{\top} \mW \mV$ where $\mU$ and $\mV$ are the left/right top $r$ singular vectors of either $\mW$, a random matrix, or $\Delta \mW$ of PEFTs. The weight matrices are taken from the query projection of the 16th layer of LLaMA2-7B.}
\label{tab:delta_w}
\end{table*}
\begin{figure*}[htbp]
    \centering
    \includegraphics[width=1.0\textwidth]{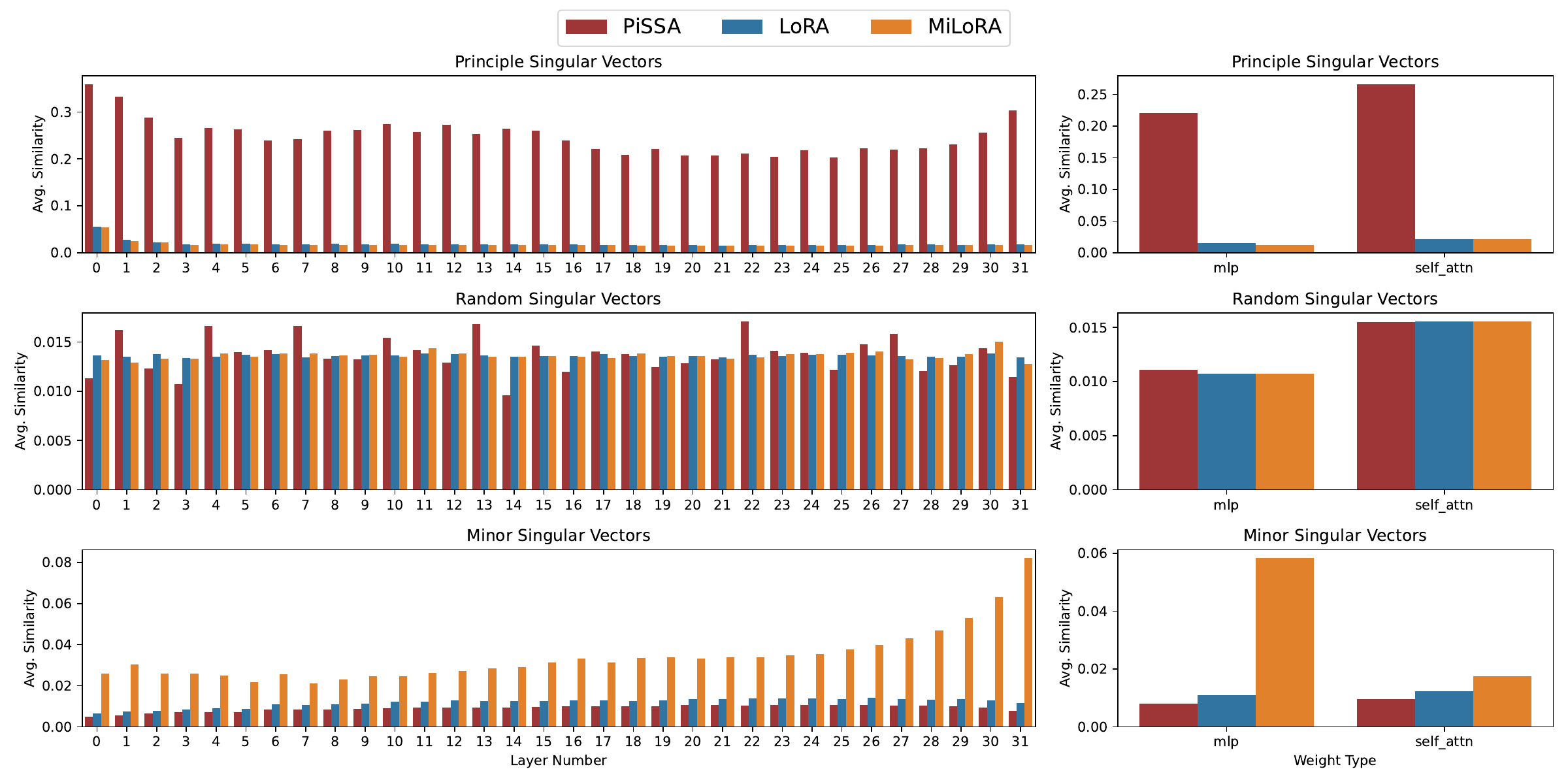}
    \caption{\textbf{Left:} The subspace similarity averaged by all modules in a layer. \textbf{Right:} The subspace similarity averaged by all layers for a specific module type.}
    \label{fig:similarities}
\end{figure*}

\subsection{How does the Matrix Update $\Delta W$ Compare to $W$?}\label{sec:subspace}

We investigate the relationship between $\Delta \mW$ and $\mW$ of different methods, where $\Delta \mW$ is the difference between the finetuned and pre-trained weights. Following the analysis method in \citet{hu2021lora}, we project $\mW$ and $\Delta \mW$ onto the $r$-dimensional subspace of $\Delta \mW$ by computing $\mU^T \mW \mV$ and $\mU^T \Delta \mW \mV$, with $\mU$/$\mV$ being the left/right top $r$ singular-vector matrix of $\Delta \mW$. As a comparison, we also compute $\mU^T \mW \mV$ by replacing $\mU$, $\mV$ with the top $r$ singular vectors of $\mW$ or a random matrix. The results are shown in Table~\ref{tab:delta_w}. Following \citet{hu2021lora}, we use the query weight in the middle layer (16th) of the model for analysis and use the Frobenius norm as the measurement of magnitude.

We make two key observations. First, $\Delta \mW$ of LoRA and \mname have a similar correlation with $\mW$, both slightly stronger than a random matrix, indicating that they amplify directions not emphasized in $\mW$. In contrast, $\Delta \mW$ of PiSSA shows a strong correlation with $\mW$. This is attributed to PiSSA's direct optimization of the principal singular components. Second, the amplification factor varies significantly across different methods. Compared to LoRA, we find the amplification factor of \mname is significantly smaller. This indicates that \mname has a reduced impact on the features which already present in $\mW$. See Table~\ref{tab:delta_w_mlp_down} of Appendix~\ref{sec:app_w} for analyzing the weight matrix of MLP down projection in the same layer, which demonstrates similar trends.

To further analyze the characteristics of $\Delta \mW$ for different methods, we measure the normalized subspace similarity \citep{hu2021lora} between the subspaces spanned by top $r$ singular vectors of $\Delta \mW$ and the subspace spanned by top $r$ singular vectors, bottom $r$ singular vectors, and a random $r$ singular vectors of $\mW$. We define the normalized similarity metric based on the Grassmann distance following \citet{hu2021lora} (See Appendix~\ref{sec:subspace_similarity} for more details) and report the results in Figure~\ref{fig:similarities}. From the left subfigure, it is clear that PiSSA learns in directions closely aligned with the top singular vectors of the pretrained weights, while LoRA and \mname do not significantly optimize in these directions. Specifically, \mname focuses on optimizing directions associated with the bottom singular vectors of the pretrained weights. The right subfigure further shows that weight matrices in both MLP and self attention modules of \mname learn in directions more closely aligned with the bottom singular vectors, especially the MLP modules.

\subsection{\mname Forgets Less than Baselines}\label{sec:forget}

\begin{table}[t]
\centering
\resizebox{0.8\columnwidth}{!}
{
\begin{tabular}{ccccc}
\toprule
Method & LoRA & PiSSA & \mname \\
\midrule
Forgetting loss & 3.24 & 6.07 & 2.54 \\
\bottomrule
\end{tabular}}
\vspace{10pt} 
\caption{Forgetting loss of various PEFT methods. }
\label{tab:forgetting}
\vspace{-8pt}
\end{table}

A hypothesis explanation for why \mname outperforms baselines is that pre-trained knowledge is more retained. To test this hypothesis, we follow \citet{kalajdzievski2024scaling} and use cross-entropy as the metric for measuring forgetting. This is the usual next token prediction loss used when training LLMs, except that the target next token is replaced by the distribution predicted by the pre-trained base model. We evaluate the forgetting metric on the LLaMA2-7B model finetuned in math reasoning using the WikiText-103 test dataset following \citet{kalajdzievski2024scaling}. As shown in Table~\ref{tab:forgetting}, \mname exhibits the lowest forgetting loss, which is consistent with our hypothesis that \mname makes the least modification to pre-trained knowledge. In contrast, the loss of PiSSA is significantly higher than that of the other methods.

\subsection{Comparison Between \mname and PiSSA} \label{sec:pissa}

\begin{table}[t]
\centering
\resizebox{1.0\columnwidth}{!}
{
\begin{tabular}{ccccc}
\toprule
Hyp. Setup & PEFT & GSM8K & MATH & Avg. \\
\midrule
\multirow{3}{*}{PiSSA} & LoRA & 41.5 & 5.8 & 23.6  \\
& PiSSA & \textbf{51.3} & \textbf{7.6} & \textbf{29.4}  \\
& \mname & 40.0 & 5.2 & 22.6  \\ \midrule
 \multirow{3}{*}{LLM-Adapters} & LoRA & 56.6 & 10.8 & 33.7 \\
& PiSSA &51.3 & 10.4 & 30.8 \\
& \mname & \textbf{58.6} & \textbf{11.6} & \textbf{35.1} \\
\bottomrule
\end{tabular}
}
\vspace{10pt} 
\caption{Math reasoning evaluation results for LLaMA2-7B with PiSSA hyperparameters and our (LLM-Adapters) hyperparameters.}
\label{tab:pissa}

\end{table}

Concurrent with our research, \citet{meng2024pissa} have recently introduced a low-rank adaptation method called PiSSA. PiSSA shares a similar framework as \mname, but adapts the principal singular components. We argue that PiSSA and \mname are fundamentally different. 

\noindent \textit{\textbf{Motivation:}} PiSSA is designed to approximate full finetuning by adapting the principal singular components, which are believed to capture the essence of the weight matrices. In contrast, our method \mname aims to adapt to new tasks \textbf{while maximally retaining the base model's knowledge}. To achieve this, we instead finetune the minor singular components of the weight matrices, which are less important for the pretrained weight matrices.

\noindent \textit{\textbf{Performance:}} The PiSSA paper claims that PiSSA outperforms both LoRA and adapting minor singular components. However, our investigation suggests that this claim is likely due to the specific hyperparameters used in their experiments. To verify this, we replicate their experimental setup using the same MetaMathQA 100K dataset and compare the performance of PiSSA, LoRA, and \mname under both PiSSA's hyperparameters and our own. Specifically, we choose a rank of 128 where PiSSA obtains the best performance in their paper and finetunes for 1 epoch for all setups following PiSSA paper. The results are summarized in Table \ref{tab:pissa}. 

We have \textbf{several observations}. \textbf{First}, under the PiSSA hyperparameter setup, PiSSA does indeed outperform both LoRA and \mname, consistent with the claims made in the original PiSSA paper. \textbf{Second}, we find that all methods perform better in our hyperparameter setup, suggesting that our hyperparameter configuration is more effective. PiSSA configuration of hyperparameters has a combination of a small learning rate (2e-5) and a large batch size (128). This could potentially result in slow learning speed and sub-optimal performance for common PEFTs, as the optimal learning rate for common PEFTs is generally much higher than that for full finetuning \citep{lialin2023scaling,biderman2024lora}. \textbf{Third}, in our hyperparameter setup, \mname performs the best, outperforming LoRA and PiSSA by 1.4 and 4.3 average scores, respectively. In summary, when using an appropriate hyperparameter configuration, our proposed \mname method achieves markedly superior performance compared to PiSSA.

\section{Related Work}

\paragraph{Parameter-Efficient Finetuning of LLMs.}
After pretraining, LLMs are finetuned with instruction data or human preference data to adapt to different downstream tasks \citep{ouyang2022instructgpt,yu2023metamath,cui2023ultrafeedback}. Parameter-efficient finetuning \citep[PEFT]{lialin2023scaling} explores effective approaches to reduce the computational resources required during finetuning while managing to match the performance of full finetuning.
Previous PEFT methods can be grouped into three lines: adapter-, LoRA- and prompt-based methods. Adapter-based methods insert additional trainable feedforward submodules into each Transformer layer \citep{houlsby19adapter,pfeiffer2021adapterfusion,pfeiffer2021unks}. However, they introduce additional computational costs during inference. LoRA-based methods \citep{hu2021lora,liu2024dora,zhang2023adalora,wang-etal-2024-lora-flow} model the changes of selected linear weights as low-rank matrices. During finetuning, given a linear matrix, only two low-rank matrices are optimized while the pretrained weight matrix is frozen. Prompt-based methods \citep{lester2021power,li2021prefix,liu2022ptuning} add additional soft prompts to the input tokens. During training, only soft prompts are updated while the pretrained model parameters are fixed. Among these PEFT methods, the LoRA-based approaches are widely used for LLM fine-tuning because they are easy to implement and do not introduce computational overhead during inference.

\paragraph{LoRA and Its Variants}
LoRA \citep{hu2021lora} reparameterizes the weight update with two trainable low-rank matrices, while freezing the pretrained weights. With this lightweight decomposition, LoRA reduces storage and task-switching overhead by sharing the pretrained models across multiple tasks. Since then, researchers have explored and proposed different LoRA variants for PEFT.
AdaLoRA \citep{zhang2023adalora} and ALoRA \citep{liu2024alora}, adaptively determine the rank of LoRA module in each weight matrix according to the importance score. 
The rsLoRA \citep{kalajdzievski2023rslora} modifies the LoRA with the appropriate scaling factor to improve the performance of large ranks.
The DoRA \citep{liu2024dora} method decomposes the pretrained weight into the magnitude and directional components, then finetunes both for better performance. Concurrent with our work, PISSA \citep{meng2024pissa} and LoRA-GA \cite{wang2024loraga} propose to better approximate full finetuning by only updating the principal singular components or aligning the gradients of low-rank updates to that of full finetuning. 

\section{Conclusion}
In this paper, we introduce \mname, a simple yet effective low-rank adaption method for LLMs. \mname effectively learns on finetuning tasks while better preserving the pretrained knowledge by adapting the minor singular components of pretrained weight matrices. We investigate the effectiveness of \mname on a wide range of LLM evaluation benchmarks, including commonsense reasoning, math reasoning and instruction-following, and vision-language model evaluation benchmarks. Experiment results demonstrate that \mname consistently outperforms LoRA and PiSSA without sacrificing training or inference efficiency. We hope that our work will inspire future research on parameter-efficient finetuning of LLMs. 

\section*{Limitations} \label{app:limit}

Due to limited computational resources, we primarily evaluate the effectiveness of \mname on LLaMA-family LLMs and LLaVA-1.5, focusing on tasks such as commonsense reasoning, math reasoning, and instruction following, in line with prior work~\cite{liu2024dora,meng2024pissa,wu2024reft}. We examine and compare \mname with baselines using the hyperparameter configurations provided in the previous works \cite{hu2023llmadapters,liu2024visual} instead of an exhaustive hyperparameter search for each task. We leave the exploration of \mname on other tasks and other LLMs like Mistral and Gemma as future work.

\section*{Acknowledgements}

This project was supported by National Natural Science Foundation of China (No. 62306132, No. 62106138) and Guangdong Basic and Applied Basic Research Foundation. We thank the anonymous reviewers for their insightful feedback on this work.

\bibliography{custom}

\appendix

\section{Detailed Experiment Setups}\label{sec:exp_setup}

\subsection{Similarity Metric Between Subspaces}\label{sec:subspace_similarity}
Following \citet{hu2021lora}, we use the measure $\phi(\mA, \mB)=\frac{\left\|\mA^{ \top} \mB\right\|_F^2}{r}$ to measure the similarity between two column orthonormal matrices $\mA,\mB\in \mathbb{R}^{d \times r}$, where $F$ donotes the Frobenius norm. The value of \(\phi(\mA, \mB)\) ranges from 0 to 1, where 1 indicates complete overlap and 0 indicates complete separation.

Given two matrices $\Delta \mW$ and $\mW$, we extract $r$ left singular vectors from each to form the subspace matrices for them, which are denoted as $\Delta \mW_s$ and $\mW_s$ , and then the subspace similarity is computed using $\phi(\Delta \mW_s,\mW_s)$.

\subsection{Our Hyperparameter Setup for LLM}\label{sec:app_hy}

Table~\ref{tab:our_hyper} shows our detailed hyperparameters. This setup follows LLM-adapters \citep{hu2023llmadapters}, therefore we denote it as LLM-adapters setup.

\begin{table}[tbh]
\centering
\resizebox{1.0\columnwidth}{!}{
\begin{tabular}{cccc}
\toprule
\textbf{Hyperparameters} & \textbf{ComR} & \textbf{MathR} & \textbf{InsF} \\
\midrule
Rank r & 32 & 64 & 32 \\
$\alpha$ of LoRA & 64 & 128 & 64 \\
$\alpha$ of PiSSA/\mname & 32 & 64 & 32 \\
Dropout & \multicolumn{3}{c}{0.05} \\
Optimizer & \multicolumn{3}{c}{AdamW} \\
LR & \multicolumn{3}{c}{3e-4} \\
LR Scheduler & \multicolumn{3}{c}{Linear} \\
Batch size & \multicolumn{3}{c}{16} \\
Warmup Steps & \multicolumn{3}{c}{100} \\
Epochs & \multicolumn{3}{c}{3} \\
Placement &  \multicolumn{3}{c}{query, key, value, MLP up, MLP down} \\
\bottomrule
\end{tabular}}
\caption{Our hyperparameter configuration on the commonsense reasoning (ComR), math reasoning (MathR) and instruction-following (InsF) tasks.}
\label{tab:our_hyper}
\end{table}

\begin{table}[t]
\centering
\resizebox{1.0\columnwidth}{!}{
\begin{tabular}{cc}
\toprule
\multicolumn{2}{c}{\textbf{Visual instruction tuning hyperparameters}} \\
\midrule
Rank r & 128 \\
$\alpha$ of LoRA & 256 \\
$\alpha$ of PiSSA/\mname & 128 \\
Dropout & 0.05 \\
Optimizer & AdamW \\
LR & 2e-4 \\
LR Scheduler & cosine \\
Batch Size & 16 \\
Warmup Ratio & 0.03 \\
Epochs & 1 \\
Placement & query, key, value, output, gate, MLP up, MLP down \\
\bottomrule
\end{tabular}}
\caption{Our hyperparameter configurations applied in visual instruction tuning tasks.}
\label{tab:vis_ins_hyper}
\end{table}

\subsection{PiSSA Hyperparameter Setup for LLM} \label{sec:app_pissa}

Table~\ref{tab:pissa_hyper} shows the detailed hyperparameters from the PiSSA paper, which we denoted as PiSSA hyperparameter setup.

\begin{table}[t]
\centering
\resizebox{1.0\columnwidth}{!}{
\begin{tabular}{cc}
\toprule
\multicolumn{2}{c}{\textbf{PiSSA hyperparameters}} \\
\midrule
$\alpha$ & Same as rank r \\
Dropout & 0.0 \\
Optimizer & AdamW \\
LR & 2e-5 \\
LR Scheduler & cosine \\
Batch Size & 128 \\
Warmup Ratio & 0.03 \\
Epochs & 1 \\
Placement & query, key, value, output, gate, MLP up, MLP down \\
\bottomrule
\end{tabular}}
\caption{The detailed hyperparameter configurations used by PiSSA\cite{meng2024pissa}.}
\label{tab:pissa_hyper}
\end{table}

\subsection{Hyperparameter Setup for Vision-Language Model}
\label{sec:more_imple_details}

For hyperparameters of finetuning the vision-language model, we follow the LoRA configuration provided by \citet{liu2024visual} without tuning for all PEFT methods for a fair comparison. Table \ref{tab:vis_ins_hyper} shows the details.

\section{More Experiment Results}
\subsection{Experiments on Qwen2.5-7B}
\label{sec:qwen}

\begin{table}[t]
\centering
\begin{tabular}{cccc}
\toprule
Method & GSM8K & MATH &Avg. \\
\midrule
LoRA & 81.8 & \textbf{49.3} & 65.5 \\
PiSSA & 81.7 & 45.4 & 63.6 \\
\mname & \textbf{85.5} & 48.7 & \textbf{67.1} \\
\bottomrule
\end{tabular}
\caption{Math reasoning evaluation results for Qwen2.5-7B on GSM8K and MATH.}
\label{tab:qwen}
\end{table}

To enhance the credibility of our experiments, we evaluate MiLoRA using Qwen2.5-7B on the MetaMathQA-100K dataset. As shown in Table~\ref{tab:qwen}, MiLoRA outperforms LoRA and PiSSA by an average of 1.6 and 3.5 EM scores, respectively, demonstrating its effectiveness on the Qwen backbone. Additionally, we conduct experiments on LLMs such as LLaMA2 and LLaMA3 (Section~\ref{exp:llm}), as well as vision-language models like LLaVA-1.5 (Section~\ref{sec:vl_exp}). Collectively, these results confirm the effectiveness of MiLoRA across diverse model architectures.
\subsection{Complementary Weighet Metrics Analysis}
\label{sec:app_w}

\begin{table}[ht]
\centering
\resizebox{1.0\columnwidth}{!}{
\begin{tabular}{r|cc|ccc}
\toprule
$\left\|\mW\right\|_F=123.0$ & $\mW$ &  Random & LoRA & PiSSA & \mname \\
\midrule
$\left\|\mU^{\top} \mW \mV \right\|_F=$ & 33.34 & 1.17 & 1.32 & 6.98 & 1.29 \\
\midrule
$\left\|\mU^{\top} \Delta \mW \mV \right\|_F=$ & - & - & 77.02 & 74.34 & 56.61 \\
\midrule
Ampliﬁcation factor & - & - & 58.35 & 10.65 & 43.88 \\
\bottomrule
\end{tabular}}
\caption{The Frobenius norm of $\mU^{\top} \mW \mV$ where $\mU$ and $\mV$ are the left/right top $r$ singular vectors of either $\mW$, a random matrix, or $\Delta \mW$ of PEFTs. The weight matrices are taken from the MLP down projection of the 16th layer of LLaMA2-7B.}
\label{tab:delta_w_mlp_down}
\end{table}

We present additional results about the relationship between $\Delta \mW$ and $\mW$ in Table~\ref{tab:delta_w_mlp_down}, in which the weight matrices are taken from the MLP down projection of the 16th layer of LLaMA2-7B.

\section{Broader Impacts} 

The \mname method enhances model performance with lower training costs. Teams or users with limited computational resources can finetune large models using \mname, promoting the broader application of large models across diverse groups.
However, the PEFT methods could potentially be exploited to finetune models for malicious purposes. It is essential to investigate rules and strategies to prevent the misuse of PEFT methods.

\end{document}